\documentclass{isprs} 
\usepackage{subfigure}
\usepackage{setspace}
\usepackage{geometry} 
\usepackage{epstopdf}
\usepackage[labelsep=period]{caption}  
\usepackage[british]{babel} 
\usepackage[hang]{footmisc}
\usepackage{color}
\usepackage{colortbl}
\usepackage{tabularx,booktabs}

\begin{document}

\title{3D Gaussian Splatting aided Localization for Large and Complex Indoor-Environments}
\date{}


\author{***** (for review, names must be rendered anonymous)}
\author{Vincent Ress\textsuperscript{1}, Jonas Meyer\textsuperscript{1,2}, Wei Zhang\textsuperscript{1}, David Skuddis\textsuperscript{1}, Uwe Soergel\textsuperscript{1}, Norbert Haala\textsuperscript{1}} 

\address{\textsuperscript{1 }Institute for Photogrammetry and Geoinformatics, University of Stuttgart, Germany -  forename.lastname@ifp.uni-stuttgart.de\\
\textsuperscript{2 }Institute of Geomatics, University of Applied Sciences and Arts Northwestern Switzerland, Switzerland - jonas.meyer@fhnw.ch\\
}

\abstract{
The field of visual localization has been researched for several decades and has meanwhile found many practical applications. Despite the strong progress in this field, there are still challenging situations in which established methods fail. We present an approach to significantly improve the accuracy and reliability of established visual localization methods by adding rendered images. In detail, we first use a modern visual SLAM approach that provides a 3D Gaussian Splatting (3DGS) based map to create reference data. We demonstrate that enriching reference data with images rendered from 3DGS at randomly sampled poses significantly improves the performance of both geometry-based visual localization and Scene Coordinate Regression (SCR) methods. Through comprehensive evaluation in a large industrial environment, we analyze the performance impact of incorporating these additional rendered views.
}

\keywords{Dense Reconstruction, Indoor Localization, SLAM, Autonomous Indoor Construction}

\maketitle


 
\sloppy

\section{Introduction}\label{Vincent}
Dense 3D scene reconstruction from imagery remains a core challenge in computer vision, robotics, and photogrammetry. Recent breakthroughs in deep learning, including 3D Gaussian Splatting (3DGS) \cite{kerbl2024hierarchical}, have significantly advanced both the performance and visual quality of the reconstruction. Within our work, we focus on 3D mapping of complex, large-scale indoor environments such as construction sites and factory halls. This initiative is driven by a project within the Cluster of Excellence Integrative Computational Design and Construction for Architecture (IntCDC) at the University of Stuttgart, which aims to enable autonomous indoor construction for new or preexisting buildings (IntCDC, 2024a).  

Typical construction tasks, including material handling and element assembly, require highly accurate mapping approaches to enable precise localization of both building components and the construction robots. Image-based localization methods are particularly valuable due to the widespread availability and low cost of cameras, which are now standard equipment on most modern robots. This universal availability means localization algorithms can be deployed across diverse environments and robotic platforms with minimal modifications. In well-observed areas with rich texture, current image-based localization algorithms can achieve exceptional accuracy in determining both position and orientation. However, these methods' performance deteriorates significantly when confronted with substantial perspective differences or large baselines between reference data and images being processed for localization. Within the scope of this work, we present and evaluate a solution that is intended to reduce the influence of perspective differences in particular by increasing the number of reference data in synthesizing novel views.

\begin{figure}[hbt!]
\centering
\includegraphics[width=0.49\textwidth]{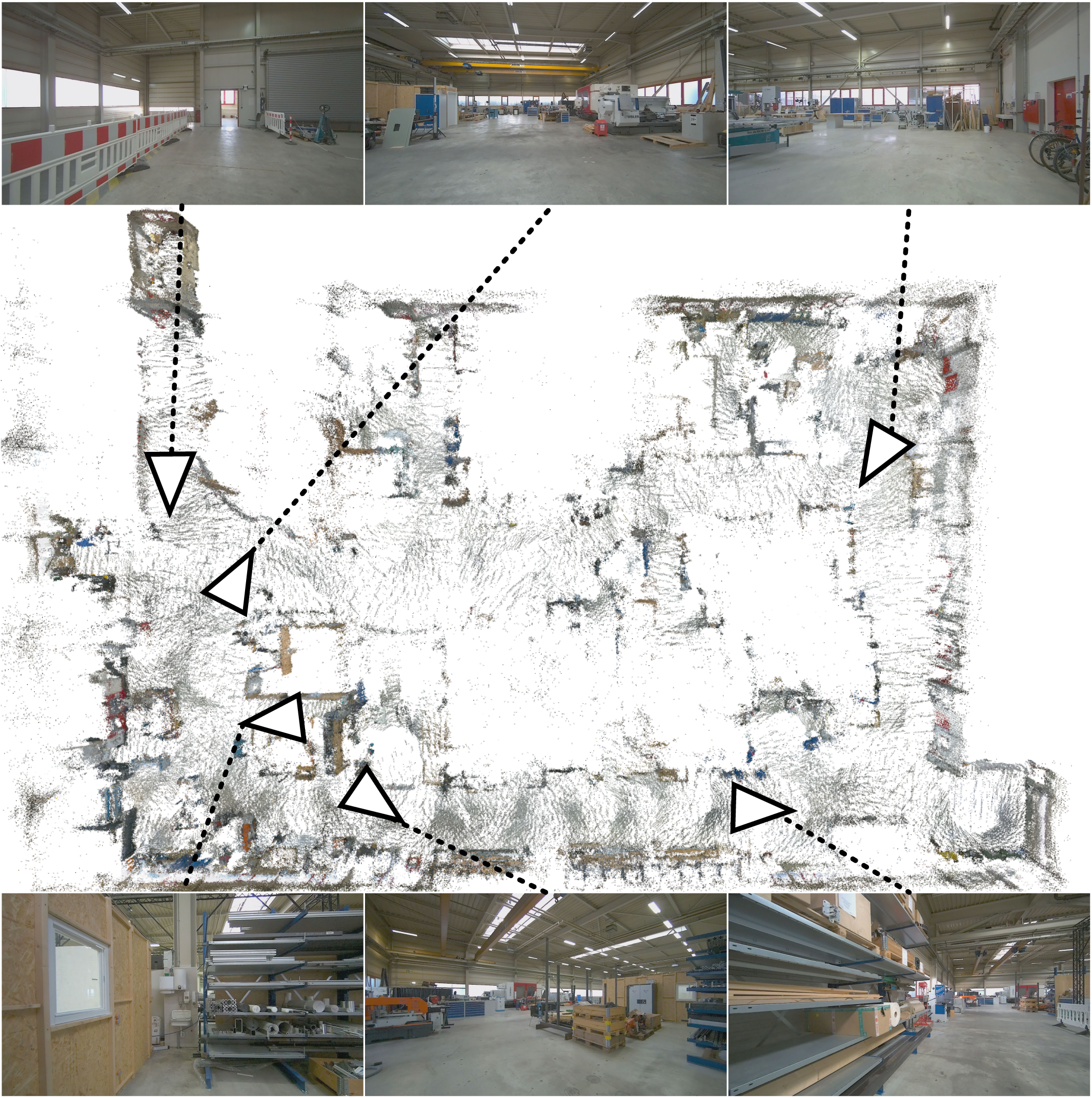}
\caption{Construction hall used as test scenario}
\label{fig:IntCDC_Construction_Hall_Overview}
\end{figure}

An essential step in generating novel views is the collection of data and mapping of the environment. While Terrestrial Laser Scanning (TLS) has traditionally been the standard for high accurate 3D mapping in indoor and Global Navigation Satellite Systems (GNSS)-denied environments, it also comes with drawbacks - namely long processing and acquisition times, including the need for manual environment preparation. Mobile mapping systems utilizing Simultaneous Localization and Mapping (SLAM)-based approaches have emerged as a more flexible and efficient alternative. However, despite offering real-time capabilities, visual SLAM algorithms typically produce less accurate and sparse maps, with particular vulnerability to scaling issues in large open spaces \cite{ress2024slam}.

To support the project’s long-term goal of enabling high-frequency, fully autonomous robotic surveys of construction sites, this study concentrates on efficient and adaptable SLAM-based acquisition algorithms. Additionally, a photogrammetric reference based on Structure-from-Motion (SfM) was created to benchmark the accuracy of image-based mapping algorithms. Using the computed camera poses and point clouds, a 3D Gaussian Splatting (3DGS) environment is initialized and optimized (cf. Chapter \ref{DenseSLAM}). Research by \cite{ress2024slam} has demonstrated that this representation method is particularly effective in visualizing large and structurally complex indoor environments. The final stage demonstrates the viability of the captured 3D representation for localization purposes. The dense Gaussian Splatting environment is used to generate rendered RGB images, which then provide additional training data for the DSAC* Network \cite{brachmann2021visual}. This network architecture was selected for its state-of-the-art performance in RGB image localization, enabling scene coordinate estimation followed by a RANSAC-based camera pose computation for images to be localized. For comparison, we also implemented a geometry-based visual localization method that employs a global image descriptor to identify the most similar images with known poses, followed by a local bundle adjustment to query the pose of the input image (cf. Chapter~\ref{Visual Localization}).

\begin{figure*}[hbt!]
\centering
\includegraphics[width=0.98\textwidth]{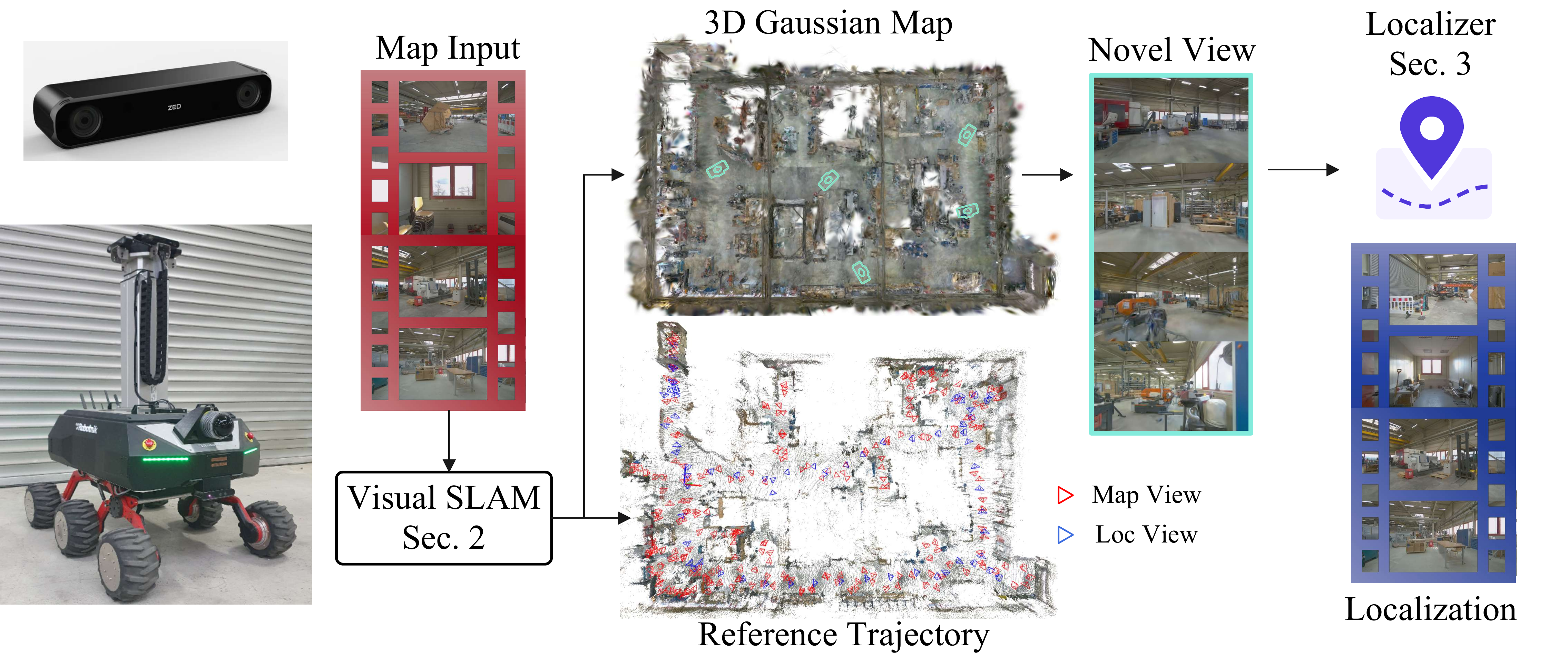}
\caption{Overview of our pipeline: first stereo inputs are processed by dense visual SLAM to provide camera poses and initial scene structure, which are refined through 3D Gaussian Splatting to create a high-quality scene representation. This enables rendering of novel views to complement real images for robust visual localization in complex indoor environments.}
\label{fig:overview}
\end{figure*}

\section{Dense SLAM for 3D Gaussian mapping of a construction hall}\label{DenseSLAM}
As illustrated in the overview of our pipeline (Fig.~\ref{fig:overview}), we introduce a dense visual SLAM pipeline for 3D Gaussian mapping of a construction hall, serving as the foundation for robust visual localization. This section is organized as follows: We first review related work in dense mapping and 3D scene representation (Sec.~\ref{RelatedWork}). We then describe our sensor system and data acquisition setup (Sec.~\ref{SensorSystem}), followed by details of our visual SLAM and 3D Gaussian optimization pipeline (Sec.~\ref{VisualSLAM}). Next, we outline our photogrammetric reference to evaluate the accuracy of the visual SLAM pipeline (Sec.~\ref{PhotogrammetricReferenceJonas}). Finally, we evaluate the quality of our Gaussian mapping through quantitative and qualitative comparisons (Sec.~\ref{Evaluation}).

\subsection{Related Work}\label{RelatedWork}
Dense visual SLAM remains fundamental for autonomous navigation and 3D mapping in robotics and photogrammetry. While traditional SLAM systems like ORB-SLAM \cite{mur2015orb} use sparse feature matching for pose tracking, they often struggle in complex environments with low texture or repetitive patterns. Dense SLAM approaches such as LSD-SLAM \cite{engel2014lsd} and DSO \cite{engel2017direct} address this by leveraging dense image information, improving robustness in challenging scenarios. More recently, learning-based systems like DROID-SLAM \cite{teed2021droid} have emerged, using neural networks to achieve more accurate pose estimation and complete 3D scene representations. However, the resulting point clouds and meshes have limitations - point clouds often lack fine details and textures, while high-resolution meshes require significant storage. 3D Gaussian Splatting (3DGS) \cite{kerbl2024hierarchical} offers a promising alternative by providing a compact yet expressive scene representation that maintains both visual fidelity and geometric accuracy across viewpoints. By integrating Gaussian Splatting with dense SLAM, as demonstrated in HI-SLAM2 \cite{zhang2024hislam2}, we can produce detailed 3D representations of complex indoor environments like construction halls. This approach not only improves the quality of the reconstructed scene but also enables the rendering of novel views to enhance mapping coverage and robust visual localization.

\subsection{Sensor System and Data Acquisition} \label{SensorSystem}

A section of the Large-Scale Construction Robotics Laboratory (LCRL) within the Cluster of Excellence IntCDC was selected as the test environment \cite{IntCDC_LCR}. The facility comprises a large construction hall that houses multiple robotic building part prefabrication plants, instructor workspaces, material storage areas, and traditional fabrication tools. The environment features both expansive open spaces and narrow corridors (cf. Fig.~\ref{fig:IntCDC_Construction_Hall_Overview}). Within the recorded area, various structures constructed from materials including concrete, steel, and wood are present.

The experimental setup utilizes a 6-wheeled robotic platform that provides essential functionalities including power supply, computational resources, and mobility capabilities (cf. Fig. \ref{fig:overview}). Data collection was performed using a camera head equipped with four ZED X RGB-D cameras. To optimize processing performance, only the forward-facing camera was utilized (1080p@10fps) and analyzed in the subsequent mapping procedures. To create a comprehensive reference map, the robot traversed all accessible areas within its physical dimensions. Where feasible, paths were covered in both directions. The localization data was collected during a separate, independent run to ensure translational and rotational deviations from the reference poses and allow for a thorough evaluation of localization performance. The total mapped area encompassed approximately $2500m^2$. 

\subsection{Visual SLAM}\label{VisualSLAM}
The goal of the Gaussian map is to provide a compact yet expressive scene representation that maintains high visual quality and geometric accuracy across different viewpoints as shown in Fig.~\ref{fig:overview}. Inspired by~\cite{ress2024slam}, we first employ a dense visual SLAM~\cite{teed2021droid} with stereo camera input to obtain camera poses and reconstruct a point cloud map of the observed scene, providing an initial sparse geometric structure of the environment. This point cloud serves as initialization for the 3D Gaussian primitives.

To create the final map, we optimize the parameters of the Gaussian primitives (position, orientation, scale, opacity, and color) through multiple loss functions. The primary objective is achieved through a photometric loss that measures color consistency between rendered and observed images. To enhance geometric accuracy~\cite{zhang2024hi}, we incorporate stereo depth measurements as a geometric constraint, enforcing consistency between the rendered depths and the estimated stereo depths. Additionally, to handle varying exposures across different viewpoints~\cite{kerbl2024hierarchical,zhang2024hislam2}, we estimate a 3×4 affine transformation matrix for each training view to compensate for illumination changes. To further improve the map quality, we incorporate the anti-aliasing filter proposed by~\cite{yu2024mip}. In mobile robotics applications, varying observation distances result in changing sampling rates, which can lead to rendering artifacts. The filter effectively mitigates these by regularizing the size of Gaussian primitives. Through iterative optimization, these combined techniques guide the refinement of Gaussian primitives to achieve accurate geometric and appearance representation.

By leveraging our Gaussian map, we can generate highly detailed color and depth images from arbitrary viewpoints. This capability allows us to generate views even for areas that were not directly captured during data acquisition. Thus to enhance view coverage, we systematically augment our dataset by sampling 25 additional poses for each keyframe, varying both position and orientation. Along the robot's driving direction (longitudinal), we sample within ±0.5m translation, while allowing larger lateral translations of ±2.0m to expand coverage perpendicular to the path. For orientation, we sample across the full ±180° yaw rotation range to enable views from all directions. While these novel views from extrapolated poses may exhibit lower visual quality or sometimes significant artifacts compared to renderings at original viewpoints (Fig.~\ref{fig:renders}), they provide valuable additional information for localization. As we demonstrate in Section~\ref{vlExperimentsResults}, our localization approaches can effectively utilize these additional rendered views to improve localization performance.

\subsection{Photogrammetric Reference}\label{PhotogrammetricReferenceJonas}
To evaluate the trajectory accuracy of our visual SLAM pipeline, we created a photogrammetric model with the SfM software Agisoft Metashape \cite{agisoftllc2023agisoft}. First, we jointly aligned all images of the mapping and the localization trajectory by using the image poses from the SLAM pipeline as pose priors. The test area lacks precise ground control points (GCPs) that could be used to minimize drift and scale issues. Therefore, we selected 26 distinctive points at strategically reasonable locations to mutually stabilize sub-loops of both trajectories. Each of these points was measured manually in around 20 to 50 images. In addition, we determined 10 reference scales distributed across the entire test area to ensure that the final model is true to scale. For the final camera optimization, the origin of the coordinate systems was set to align with the first image of the mapping trajectory, while all remaining image poses were neglected. The photogrammetric reference model has an Root Mean Square Error (RMSE) for the scale bars of $0.003 m$ and an RMSE for all re-projection errors of the manually measured tie points of $0.652 pix$.

\begin{figure}[hbt!]
\centering
\includegraphics[width=0.49\textwidth]{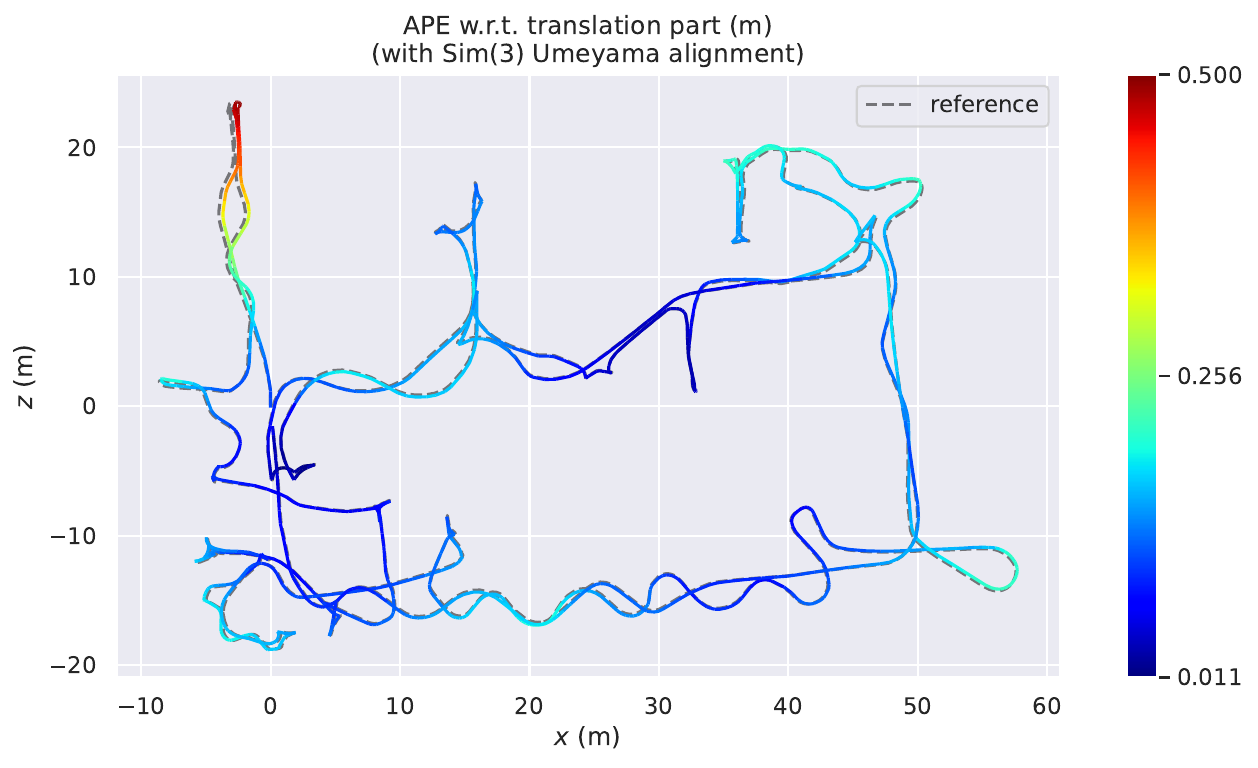}
\vspace{-7mm}
\caption{Absolute Trajectory Error of the visual SLAM pipeline relative to the photogrammetric reference.}
\label{fig:ape}
\end{figure}
We evaluate the trajectory accuracy of our visual SLAM pipeline by comparing it with the photogrammetric reference model. Fig.~\ref{fig:ape} shows the trajectory of the visual SLAM pipeline color-coded by the Absolute Trajectory Error (ATE) relative to the photogrammetric reference. The ATE measures the Euclidean distance between corresponding camera positions after aligning both trajectories using a similarity transformation. Overall, the pipeline achieves an ATE RMSE of $0.16 m$, with maximum deviations of $0.49 m$. This demonstrates the high accuracy of our approach for indoor mapping applications while maintaining real-time performance.

\subsection{Evaluation of Gaussian mapping quality }\label{Evaluation}

\begin{table}[hbt!]
\centering
\caption{Experiment configurations and map quality results.}\label{tab:configs}
\vspace{-2mm}
\begin{tabular}{c|l|l}
\toprule
Config. & Description                 & PSNR  \\
\midrule
(a)     & Color + depth Loss          & 26.14 \\
(b)     & (a) + exposure compensation & 27.03 \\
(c)     & (b) + anti-aliasing filter  & 27.55 \\
\bottomrule
\end{tabular}
\end{table}

We evaluate our Gaussian map quality through both quantitative metrics and qualitative visual comparisons between rendered and input images. Table~\ref{tab:configs} shows the systematic evaluation of three pipeline configurations using the Peak Signal-to-Noise Ratio (PSNR) metric. The baseline configuration using only color and depth loss achieves a PSNR of $26.14$. Adding exposure compensation yields a significant improvement to $27.03$ PSNR, while the final addition of the anti-aliasing filter further enhances the result to $27.55$ PSNR.

\begin{figure}[hbt!]
\centering
\includegraphics[width=0.48\textwidth]{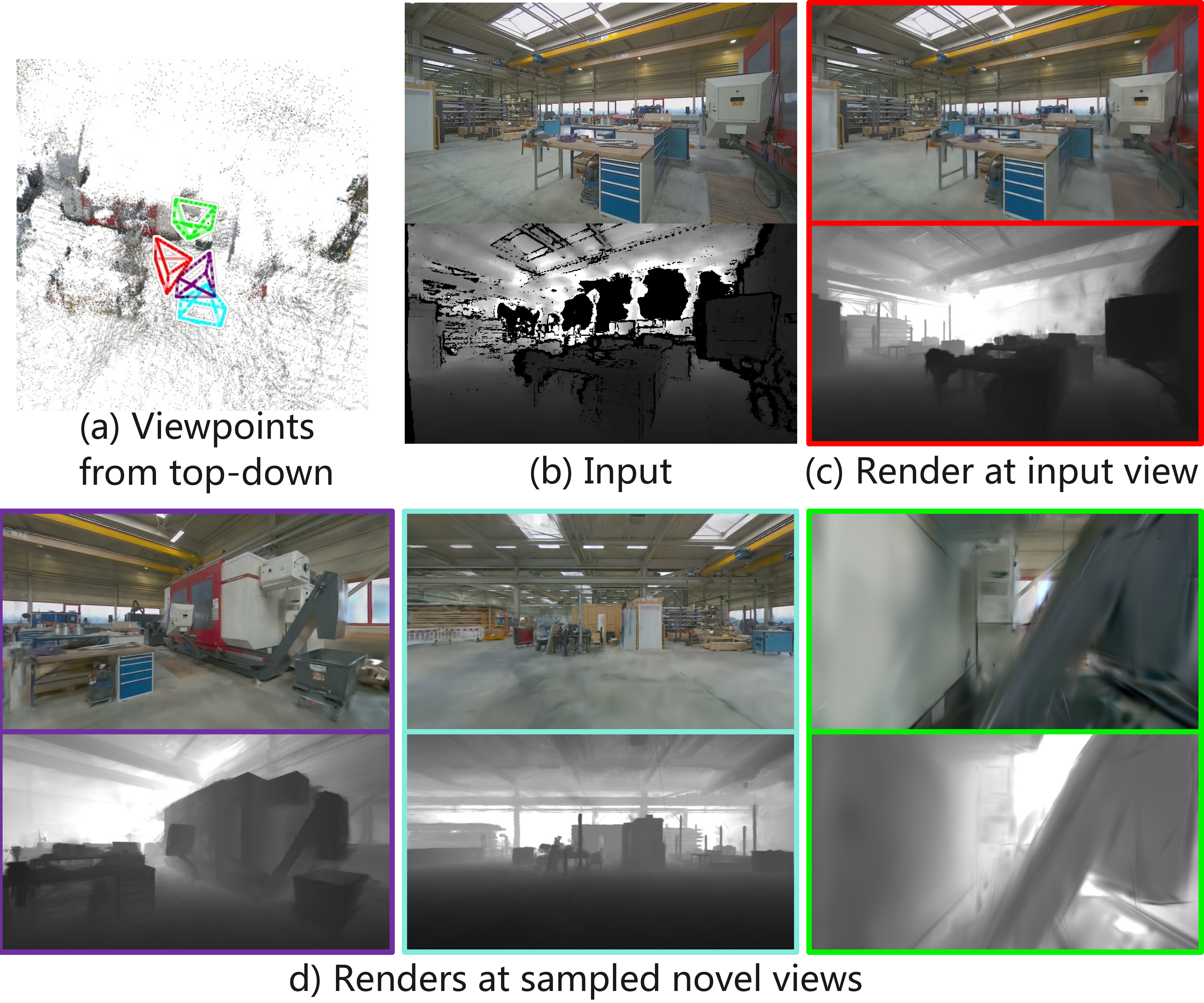}
\vspace{-5mm}
\caption{Qualitative results: (a) top-down view showing input viewpoint(red) and sampled novel viewpoints, (b) input image, (c) rendered image at input view, and (d) rendered images from sampled novel viewpoints.}
\label{fig:renders}
\end{figure}

As demonstrated in Fig.~\ref{fig:renders}b), our method produces rendered views with high visual fidelity compared to the input images. The depth images show particularly notable improvements, with enhanced geometric detail preservation at object boundaries and better clarity of background structures at greater distances. Furthermore, our improved Gaussian map enables high-quality renderings even at extrapolated poses with large translational and rotational differences from the input views as shown in Fig.~\ref{fig:renders} d). This capability proves valuable for enhancing localization performance, as demonstrated in our experimental results in Section~\ref{LocalizationApproaches}.

\section{Visual Localization}\label{Visual Localization}
In this section, we present our visual localization approaches aiming at precise robot re-localization within the previously created reference map. This section is structured as follows: First, we provide a brief overview of related work (Sec.~\ref{vlRelatedWork}). Subsequently, we describe our employed approaches in detail (Sec.~\ref{LocalizationApproaches}). Finally, we outline the conducted localization experiments and present the associated results (Sec.~\ref{vlExperimentsResults}).

\subsection{Related Work}\label{vlRelatedWork}
Visual (re-)localization describes the task of estimating a camera’s pose with respect to a known scene. This task is a core component in different fields of research such as robotics, autonomous navigation and augmented reality. However, despite great efforts in research, visual localization is still facing major challenges when strong changes in viewpoint, illumination and appearance are present. We divide visual localization methods into two main categories: matching-based and regression-based methods. Regression-based methods can directly estimate a query image's pose based on the estimated scene coordinates, offering faster inference times compared to matching-based alternatives. In contrast, matching-based methods determine pose by actively comparing query images against a reference database. While this comparison process is more computationally intensive and the execution time depends on the database size, it currently achieves superior over all pose accuracy compared to regression-based approaches.

\subsubsection{Matching based Methods}\label{Jonas}
Matching-based methods can be further distinguished between \textit{image retrieval} and \textit{geometry-based} methods.

Image retrieval methods aim at finding the \textit{k} most similar images for a query image within a reference database. This is usually done by extracting descriptors and performing nearest neighbor search in the descriptor domain. Traditionally, concepts such as Bag-of-Words (BoW) \cite{sivic2003VideoGoogletext}, Histogram of Oriented Gradients (HOG) \cite{dalal2005HOG} or Vector of Locally Aggregated Descriptors (VLAD) \cite{jegou2010Aggregatinglocaldescriptors} were employed. With the advent of deep learning, learned global image descriptors have become prevalent. Approaches such as NetVLAD \cite{arandjelovic2016NetVLADCNNarchitecture} or more recently EigenPlaces \cite{berton2023EigenPlacesTrainingViewpoint} and AnyLoc \cite{keetha2023anyloc} have advanced the state-of-the-art by using novel network architectures and training schemes. The accuracy of image retrieval methods is usually low, depending on the density of the reference images.

In contrast, geometry-based methods offer highly accurate pose estimates, in the range of a few centimeters. By employing correspondences between pixels in the query image and 3D points of the reference map, the query image's pose can be estimated via PnP/RANSAC \cite{gao2003pnp,fischler1981random}. Over the last two decades an increasing amount of approaches to estimate those correspondences has been introduced. Starting from handcrafted local features such as SIFT \cite{lowe2004DistinctiveImageFeatures} in combination with nearest-neighbor matcher, over neural-based robust feature representations \cite{detone2018SuperPointSelfSupervisedInterest,dusmanu2019D2NetTrainableCNN,zhao2023ALIKEDLighterKeypoint}, novel feature matching methods \cite{sarlin2020SuperGlueLearningFeature,lindenberger2023LightGlueLocalFeature,edstedt2024roma} towards pixel-wise dense matches \cite{sun2021LoFTRDetectorFreeLocal,chen2022ASpanFormerDetectorFreeImage}. Additionally, hierarchical approaches leveraging image retrieval to obtain pose priors for search space reduction \cite{sarlin2019coarsefineRobust} as well as different map representations such as meshes \cite{panek2022Meshloc}, NeRFs \cite{zhou2025Nerfectmatch} and 3D Gaussian Splatting \cite{botashev2024GSLoc,zhai2024splatloc} have been proposed. All these approaches aim at estimating better correspondences from more challenging images or viewpoints and consequently, estimate the query image's pose more robustly and more accurately.

\subsubsection{Regression based Methods}\label{David}
Regression based methods for visual localization can be divided into Scene Coordinate Regression (SCR) and Direct Pose Regression (DPR).

In SCR 3D scene coordinates are estimated for all pixels of a given query image. While early approaches used regression trees to perform this task \cite{shotton2013scene} \cite{valentin2015exploiting}, neural networks have become established in modern SCR approaches \cite{li2018full} \cite{dong2022visual} \cite{wang2024hscnet}.
While most approaches require an extensive training phase before the inference phase, there are also some scene-agnostic approaches \cite{yang2019sanet} \cite{revaud2024sacreg}.
The estimation of the scene coordinates is usually followed by a RANSAC \cite{fischler1981random} based pose estimation.

A series of approaches adds RANSAC-based pose estimation to the training of scene coordinates, achieving excellent results \cite{brachmann2017dsac} \cite{brachmann2021visual}. While here, despite the final training in an end-to-end fashion, is still a clear distinction between SCR and pose estimation in the internal structure, DPR methods use a single neural network to directly estimate poses for query images \cite{kendall2015posenet} \cite{wang2020atloc} \cite{wang2023robustloc}.

\subsection{Localization Approaches}\label{LocalizationApproaches}
In order to, re-localize our robot within the reference map, we use two different approaches. A Hierarchical Geometry-based Visual Localization (HGVL) (cf. Sec. \ref{hierachichalLoc}) approach and a Scene Coordinate Regression (SCR) (cf. Sec. \ref{sceneLoc}) approach. 

\subsubsection{Hierarchical Geometry-based Visual Localization}\label{hierachichalLoc}
Our hierarchical geometry-based visual localization pipeline has a comparable design as the one presented in \cite{meyer2020LongTermVL}. First, the ten most similar images to the query image are selected via image retrieval. Subsequently, a local SfM model is built via triangulation of pair-wise matched images from known poses. Finally, the pose of the query image is estimated based on the established 2D-3D correspondences using a PnP solver and a RANSAC scheme.

For image retrieval we use EigenPlaces \cite{berton2023EigenPlacesTrainingViewpoint}.
Thanks to its novel training scheme, which incorporates images from various different and also strongly differing viewpoints, EigenPlaces convinces by increased viewpoint invariance than competing image retrieval approaches. This should lead to more robust and thus more favorable constellations regarding the SfM-based scene reconstruction step. Further, we use ALIKED \cite{zhao2023ALIKEDLighterKeypoint} a local feature detection and description method and LightGlue \cite{lindenberger2023LightGlueLocalFeature} as feature matching method. The combination of ALIKED and LightGlue has proven to be very efficient and performant when matching images under challenging conditions \cite{bellavia2024IMC24}. Local scene reconstruction and registration of the query image is performed using COLMAP \cite{schoenberger2016colmap}.

In a preprocessing step, global image descriptors and local features are extracted for all reference images. Global image descriptors are stored in a database for fast nearest neighbor search in the descriptor domain. Similarly, local features are written to file to be swiftly read during the localization phase. 

\subsubsection{Scene Coordinate Regression}\label{sceneLoc}
Based on its state-of-the-art performance across benchmark datasets, including the 7 scenes \cite{shotton2013scene} and Cambridge dataset \cite{kendall2015posenet}, \cite{brachmann2021visual}`s DSAC* network was selected to evaluate SCR-based approaches. The network's architecture consists of two main components: a Fully Convolutional Neural Network (FCNN) to estimate (3D) Scene Coordinates and a subsequent Differential RANSAC (DSAC) module for pose determination.

The initial training phase uses a loss function mainly based on the re-projection error. By minimizing the projection error the underlying information of the captured environment such as gray values and geometry are implicitly incorporated in the trained network. Despite increasing the training complexity, the input was restricted to RGB images (keyframes selected during the mapping process) and their corresponding poses to enhance flexibility and reduce dependency on specific sensor systems. The network's generalization capability was improved through data augmentation, specifically by varying brightness, contrast, and orientation. However, to optimize convergence behavior, augmentation was temporarily disabled during the initial training cycles. To optimize computational efficiency, the images for both inference and training were down-scaled to 25\% of their original size, as we according to the authors \cite{brachmann2021visual} found this reduction caused no significant decrease in the localaization performance.

In the final training step, the network directly minimizes the pose error computed from the estimated scene coordinates. The end-to-end differentiability of the DSAC method (incl. a gradient-based approximation of PnP algorithm), as established by its authors \cite{brachmann2021visual}, enables iterative optimization throughout the training process.

\subsection{Experimental Evaluation and Results}\label{vlExperimentsResults}
\begin{table}[hbt!]
\centering
\caption{Localization results obtained with both methods (HGVL / SCR) on reference data sets; (a) keyframes only,  (b) keyframes and renderings from 3DGS; Localization results reported as the percentage of images localized within given translation and rotation thresholds.}

\begin{tabular}{c m{0.8cm} m{0.8cm} m{0.8cm} m{0.8cm} m{0.8cm}}
\toprule
Approach&0.5°/ 0.02m&1.5°/ 0.05m&3°/ 0.1m&5°/ 0.5m&10°/ 1m\\
\midrule
HGVL (a)& 40.9 & 77.5 & 85.7 & 90.4 & 92.4\\
HGVL (b)& 69.3 & 89.8 & 92.3 & 95.3 & 97.4\\
\midrule
SCR (a)& 0.0 & 0.0 & 1.5 & 31.6 & 56.4\\
SCR (b)& 0.0 & 0.3 & 2.3 & 66.1 & 80.4\\

\bottomrule
\end{tabular}
\label{tab:localization_results}
\end{table}

To evaluate the accuracy of our visual localization approaches, we first establish reference poses by aligning the trajectory of localization data with the reference map using the dense alignment method from ~\cite{pan2024dense}. This method achieves centimeter-level precision through global bundle adjustment, making it suitable for reference poses. We then compare the performance of our Hierarchical Geometry-based Visual Localization (HGVL) and Scene Coordinate Regression (SCR) approaches across different error thresholds. For each query image, we compute the absolute pose error as the deviation from the reference poses in both translation (meters) and rotation (degrees). The percentage of images successfully localized within various error thresholds is presented in Table~\ref{tab:localization_results}. The results demonstrate that using rendered images from additional viewpoints/perspectives substantially improves localization accuracy for both methods evaluated. Notably, even in sparsely observed regions where rendered images exhibit significant blur to the point of being unrecognizable, this degradation does not appear to significantly impact the overall performance.

\begin{figure}[hbt!]
\centering
\includegraphics[width=0.49\textwidth]{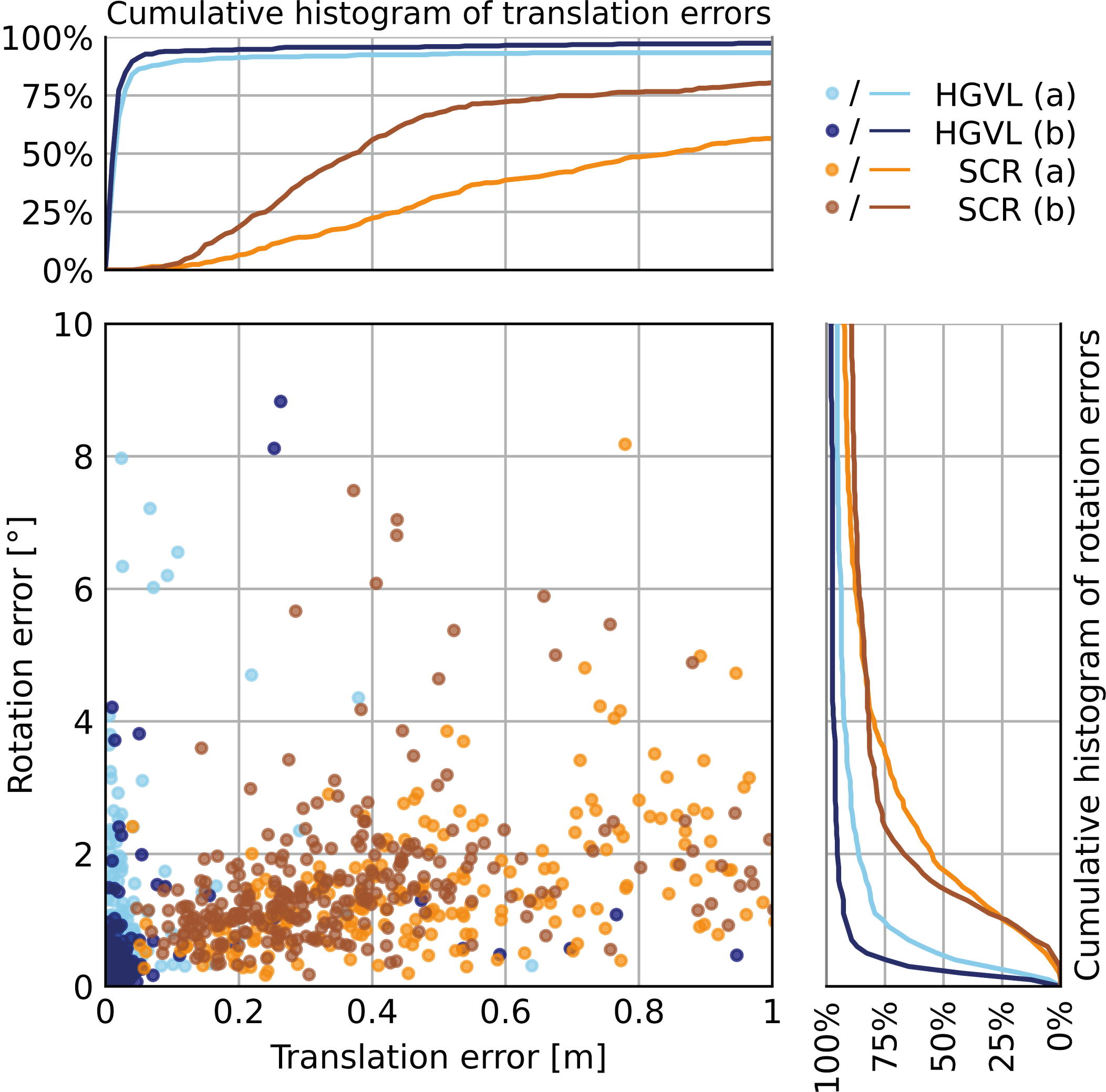}
\caption{Localization results obtained with both methods (HGVL / SCR) on reference data sets (a) keyframes only  (b) keyframes and renderings from 3DGS. Pose errors are separated into translation and rotation components.}
\label{fig:PoseErrorsLocalization}
\end{figure}

\vspace{15mm}

\begin{figure}[hbt!]
\centering
\includegraphics[width=0.49\textwidth]{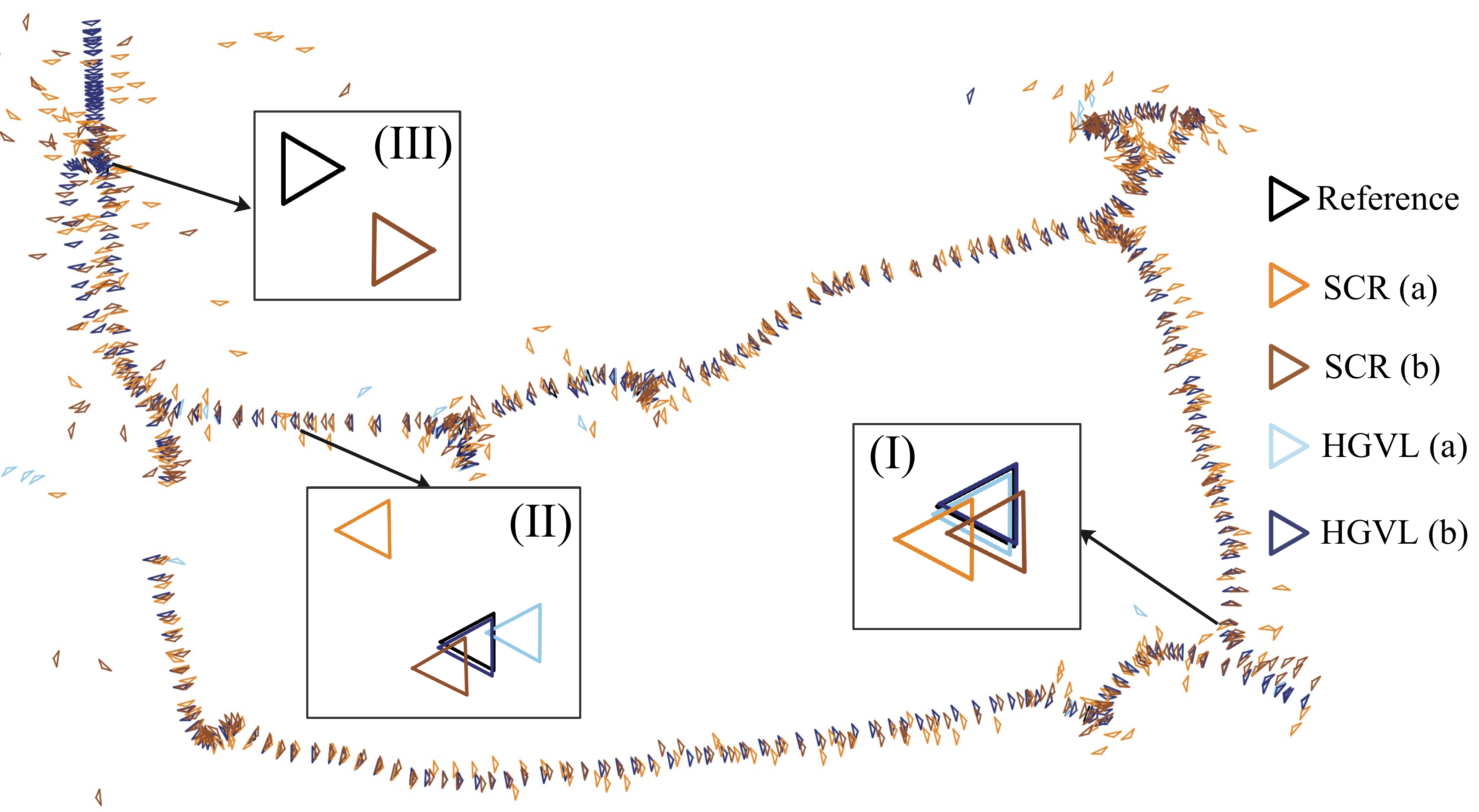}
\vspace{-6mm}
\caption{Qualitative localization results from SCR (a) in light brown to SCR (b) in dark brown and from HGVL (a) in light blue to HGVL (b) in dark blue; (a) keyframes only,  (b) keyframes and renderings from 3DGS; Three detailed examples: (I) both approaches perform well, (II) HGVL outperforms SCR, and (III) only SCR (b) succeeds.}\label{fig:locstudy}
\end{figure}

This improvement is most pronounced in the SCR-based method, which shows a $24\%$ increase in successful image localizations $(<10^{\circ}/1m)$. The enhancement is evident in all accuracy categories. Analysis of Figure~\ref{fig:PoseErrorsLocalization} and Figure~\ref{fig:locstudy} reveals that for the SCR-based method, apart from a small number of outliers, the addition of rendered training data leads to significant improvements in translation accuracy, while rotational accuracy shows only minimal change. We assume that the highly accurate poses and corresponding renderings of additional perspectives within the Gaussian environment enable a better capture of the geometry and thus a more accurate estimation of the scene coordinates. The enhanced geometric reconstruction particularly benefits depth estimation relative to the (virtual) camera, which explains the substantial reduction in translation error. Although incorporating rendered images doubled the required training time, the pose estimation inference time primarily depends on two factors: the network architecture and the number of RANSAC hypotheses. In our experiments, pose estimation for a single image required approx. $42ms$.

The HGVL-based method exhibits different characteristics. Due to its already superior baseline performance, the increase in successfully localized images is more modest at $5\%$. However, there is a notable $29\%$ improvement in high-precision localizations $(<0.5^{\circ}/0.02cm)$, indicating a significant enhancement in the method's overall accuracy. Interestingly, the HGVL-based method shows particular improvement in rotational accuracy, while maintaining its already excellent translation accuracy. This behavior may be attributed to the hierarchical image selection strategy. The inclusion of rendered images increases the likelihood of finding reference views with higher overlap to the query image. This enhanced view selection stabilizes the subsequent local bundle adjustment, ultimately improving the query image alignment and the resulting pose. We assume that especially local bundles of keyframes representing pure forward motion sequences - typically challenging for SfM approaches - benefit greatly from the rendered images (cf. Figure~\ref{fig:locstudy}). The execution times strongly depend on the number of reference images. In our setup with $29k$ rendered reference images, processing each query image took approximately $3$-$4$ seconds.

\vspace{15mm}

\section{Conclusion}\label{Vincent, Norbert}
In this work, we demonstrate that modern visual SLAM methodology can achieve highly accurate environment mapping while enabling photorealistic rendering, primarily through the integration of 3D Gaussian Splatting. Our evaluation of both scene coordinate regression and geometry-based image localization approaches shows significant improvements through the introduction of rendered novel viewpoints. By incorporating rendered imagery into reference and training datasets, we achieve dual benefits: an increased number of successfully localized images and enhanced overall accuracy for both evaluated approaches.

Since construction tasks are typically pre-planned, with robotic platform working locations known days in advance. This planning horizon enables a practical scenario where additional imagery can be rendered specifically for these locations and integrated in the reference data. In this context, the training and pre-processing times required by our approaches become negligible. 
The rapid inference times of the Scene Coordinate Regression (SCR) Network enable real-time localization during navigation. More computationally intensive but highly accurate methods like our Hierarchical Geometry-based Visual Localization (HGVL) approach can then be employed for precise final localization prior to critical operations such as drilling or assembly. However, the SCR-based approach presented here requires enhanced accuracy for construction applications.

To address this limitation, we propose leveraging the superior depth information provided by 3D Gaussian Splatting during SCR training. This approach aims to achieve two objectives: improving network convergence and enhancing the accuracy of estimated scene coordinates. Furthermore, the network can provide an initial pose estimate for the query image, enabling filtered selection of reference images and subsequent local bundle adjustment for precise alignment.

\section{Acknowledgements}
Supported by the Deutsche Forschungsgemeinschaft (DFG,
German Research Foundation) under Germany´s Excellence
Strategy – EXC 2120/1 – 390831618.

{
	\begin{spacing}{1.17}
		\normalsize
		\bibliography{ISPRSguidelines_authors} 
	\end{spacing}
}

\end{document}